\documentclass[letterpaper, 10 pt, conference]{ieeeconf}   

\IEEEoverridecommandlockouts                              

\usepackage{graphicx}
\usepackage{algorithm}
\usepackage{algorithmic}
\usepackage{mathrsfs}
\usepackage{amsmath}
\usepackage{dsfont}
\usepackage{amssymb}
\usepackage{amsfonts}
\usepackage{multirow}
\usepackage{multicol}
\usepackage{subfig}
\usepackage{epsfig}
\usepackage{subfig}
\usepackage[outline]{contour}
\usepackage{booktabs}
\DeclareMathOperator*{\argmax}{argmax}

\usepackage{hyperref}
\usepackage{cite}

\usepackage{hhline}

\begin{document}

\title{\LARGE \bf
Efficient Robotic Object Search via HIEM: Hierarchical Policy Learning with Intrinsic-Extrinsic Modeling
}

\author{Xin Ye and Yezhou Yang
                 \thanks{X. Ye and Y. Yang are with the Active Perception Group at the School of Computing, Informatics, and Decision Systems Engineering, Arizona State University, Tempe, AZ, USA, Email:
         {\tt \small  \{xinye1,  yz.yang\}@asu.edu}}

 }

\maketitle

\begin{abstract}

Despite the significant success at enabling robots with autonomous behaviors makes deep reinforcement learning a promising approach for robotic object search task, the deep reinforcement learning approach severely suffers from the nature sparse reward setting of the task. 
To tackle this challenge, we present a novel policy learning paradigm for the object search task, based on hierarchical and interpretable modeling with an intrinsic-extrinsic reward setting. More specifically, we explore the environment efficiently through a proxy low-level policy which is driven by the intrinsic rewarding sub-goals. We further learn our hierarchical policy from the efficient exploration experience where we optimize both of our high-level and low-level policies towards the extrinsic rewarding goal to perform the object search task well.
Experiments conducted on the House3D environment validate and show that the robot, trained with our model, can perform the object search task in a more optimal and interpretable way.

\end{abstract}

\section{Introduction}
Robotic object search is a task where a robot (with an on-board camera) is expected to take reasonable steps to approach a user-specified object in an unknown indoor environment. It is an essential capability for assistant robots and could serve as an enabling step for other tasks, such as the Embodied Question Answering \cite{das2018embodied}. 
Classical map-based approaches like simultaneous localization and mapping (SLAM) have been studied to address this problem for a long time, but it is also well-known that SLAM-based approaches rely heavily on sensor inputs and thus suffer from sensor noises \cite{kojima2019learn, mishkin2019benchmarking}.  
Recently, (deep) reinforcement learning (RL) has demonstrated its power at enabling robots with autonomous behaviors \cite{arulkumaran2017brief}, such as navigating over an unknown environment \cite{mirowski2016learning,zhu2017target}, manipulating objects with robot's end effectors \cite{gu2017deep,popov2017data,rajeswaran2017learning}, and motion planning \cite{chen2017socially,everett2018motion}. Under the RL setting, a robot learns the optimal behavioral policy by maximizing the expected cumulative rewards given the samples collected from its physical and/or virtual interactions with the environment. The rewards serve as the reinforcement signals for the robot to update its policy. 

\begin{figure}[t!]
\centering
\includegraphics[width=0.45\textwidth]{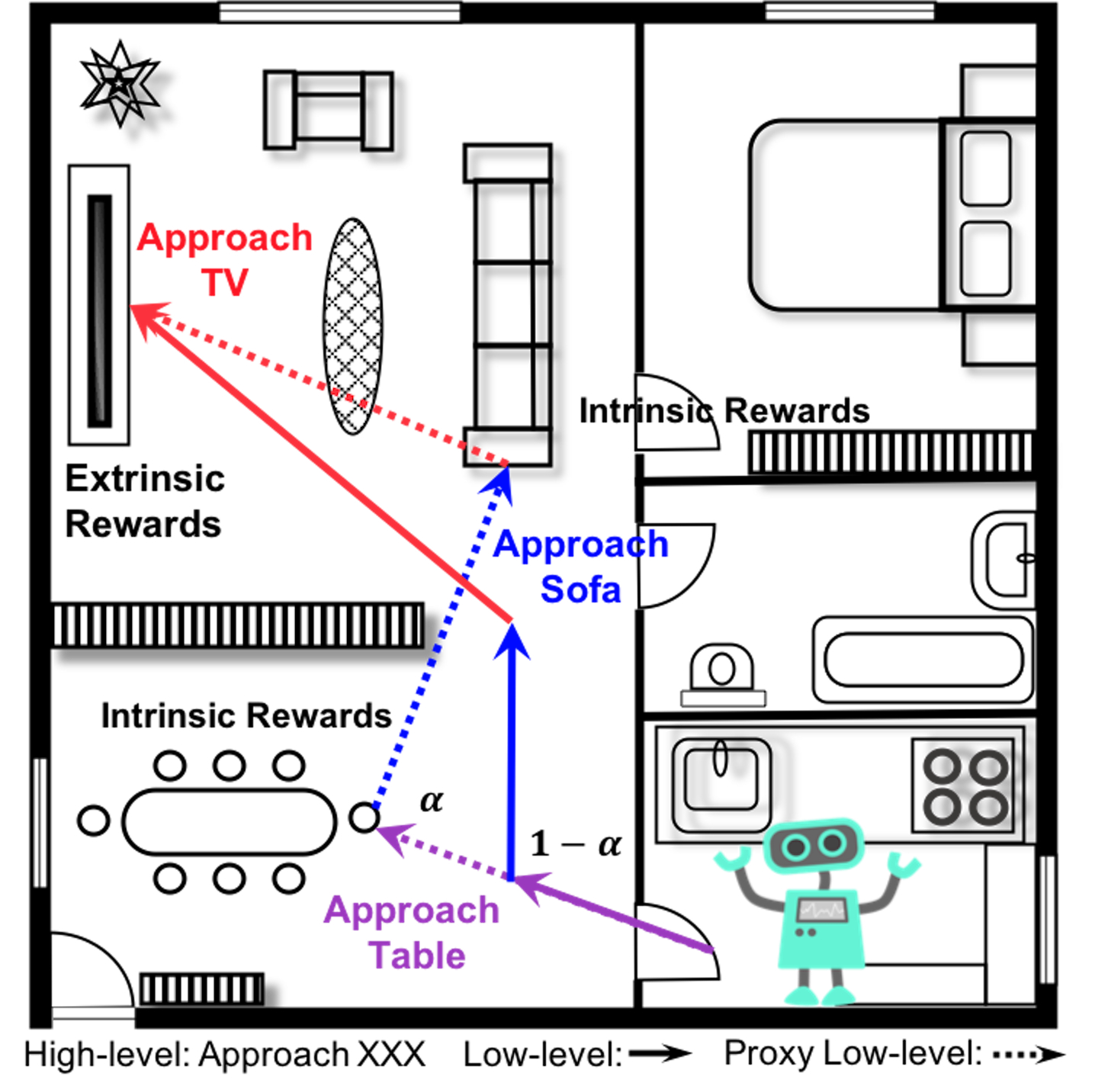}
\caption{An example of our \textit{HIEM} framework. When our high-level policy proposes a sub-goal, our proxy low-level policy is invoked with the probability of $\alpha$ to explore the environment by optimizing towards the sub-goal, and our low-level policy learned from the exploration experience is invoked with the probability of $1-\alpha$ to collaborate with the high-level policy to better achieve the goal. 
}
\vspace{-10pt}
\label{fig:concept}
\end{figure}

A pressing challenge to train a robot to perform object search with RL is the sparse reward issue, due to the fact that the environment and/or the location of the target object are typically unknown. With well-designed reward functions, such as the ones in Atari games \cite{mnih2015human}, the learned policies are shown to achieve extremely promising performance. However, it is a well-known challenge designing the reward function for the real-world applications\cite{abbeel2004apprenticeship}. Typically, for applications such as object search or target-driven visual navigation, prior research constructs the reward function in terms of the distance between the robot's current location and the object location under a strict assumption that the full information of the environment is known \cite{mousavian2019visual,wang2018look,wang2019reinforced}. For an unknown environment, a straightforward way is to set a high reward when the robot reaches the final goal state while at all other intermediate states, the reward is either zero or a small negative value \cite{zhu2017target}. More recently, \cite{ye2018active} presented a relatively denser reward function which is based on the bounding box of the target object from the robot's detection system, but the reward is still not defined among the situations where the target object is not detected. In such a sparse reward setting where the reward is only defined for a small subset of the states, the robot struggles to learn the object search policy as it is unlikely to encounter and sample the very few rewarding states without a well-designed goal-oriented exploration strategy, especially dealing with complex environments.


Hierarchical RL (HRL) paradigm is thus formulated considering its efficient strategy for exploration \cite{nachum2019does} and superiority under the sparse reward setting  \cite{kulkarni2016hierarchical,le2018hierarchical, levy2018hierarchical}. HRL aims to learn multiple layers of policies. The higher layer breaks down the task into several easier sub-tasks and proposes corresponding sub-goals for the lower layer to achieve. Typically, the sub-goals are aliases to the states that mandates the lower layer to reach, as defined in 
\cite{le2018hierarchical,nachum2018data} for tasks with low dimensional state spaces. Unfortunately, these methods are not directly applicable for the object search task in which the state observations are directly taken from the high dimensional RGB images. It is utterly difficult and seemingly impractical for the higher layer to output homogeneous images as sub-goals. On the other hand, reconstructing a concise low dimensional sub-goal space from the observation space without compromising the optimality of the learned policy demands elaborate efforts \cite{nachum2018near, dwiel2019hierarchical}.

In this paper, we put forward a novel two-layer hierarchical policy learning paradigm for the object search task. Our hierarchical policy builds on a simple yet effective and interpretable low dimensional sub-goal space. To obtain an optimal hierarchical policy given the small sub-goal space, we model the object search task with both goal dependent extrinsic rewards and sub-goal dependent intrinsic rewards. To be specific, our high-level policy plans over the sub-goal space in order to achieve the final goal by maximizing the extrinsic rewards. When a sub-goal is given following the high-level policy, a proxy low-level policy is then invoked for the robot to explore the environment. The proxy low-level policy maximizes the intrinsic rewards in order to achieve the proposed sub-goal. Meanwhile, our low-level policy learns from the exploration experience and optimizes towards the final goal. It is  invoked eventually to collaborate with our high-level policy to form an optimal hierarchical object search sequence.  Moreover, inspired by \cite{bacon2017option}, the low-level policy learns to terminate at valuable states that further improves our hierarchical object search performance. We dub our framework as \textit{HIEM}: Hierarchical policy learning with Intrinsic-Extrinsic Modeling (see Fig.~\ref{fig:concept}). We validate \textit{HIEM} with  extensive sets of experiments on the House3D \cite{wu2018building} simulation environment which contains thousands of 3D houses with a diverse set of objects and natural layouts resembling the real-world. The observed results demonstrate the efficiency and efficacy of our system over other state-of-the-art ones.


\section{Related Work}
Our work is closely related to two major research thrusts: hierarchical RL and target-driven visual navigation.

\noindent{\bf Hierarchical reinforcement learning.} Previous work has studied hierarchical reinforcement learning in many different ways. One is to come up with efficient methods to accelerate the learning process of the general hierarchical reinforcement learning scheme. As in \cite{nachum2018data}, the authors introduce an off-policy correction method. \cite{levy2017hierarchical} and \cite{levy2018hierarchical} propose to use Hindsight Experience Replay to facilitate learning at multiple time scales. Though these methods' performance are impressive, they typically assume the sub-goal space for the higher level policy is the state space. However, in the object search task, the RL system takes the image as the state representation, these methods are not directly applicable since the higher layer can hardly propose an image as a sub-goal for the lower layer to achieve. 

Other methods designate a separate sub-goal space for hierarchical reinforcement learning. For example, \cite{kulkarni2016hierarchical} defines the sub-goal space in the space of entities and relations, such as the ``reach'' relation they use for their Atari game experiment. Sub-tasks and their relations are provided as inputs in \cite{andreas2017modular} and \cite{sohn2018hierarchical}. Closer related to our work, \cite{das2018neural} adopts \emph{\{exit-room, find-room, find-object, answer\}} as the sub-goal space to learn a hierarchical policy for the Embodied Question Answering task. For the same task, \cite{gordon2018iqa} chooses \emph{\{navigate, scan, detect, manipulate, answer\}} as the possible sub-tasks, while the reinforcement learning methods are mainly applied for learning high-level policy, i.e. planning over the pre-trained or fixed sub-tasks. 

On the other side, attempts have been made to learn a set of low-level skills automatically to achieve the goal. These low-level skills are also referred to as temporal abstractions. \cite{bacon2017option} proposes the option-critic framework to autonomously discover the specified number of temporal abstractions. \cite{osa2019hierarchical} learns the temporal abstractions through advantage-weighted information maximization. \cite{nachum2018near} addresses the sub-goal representation learning problem. With the learned representation, their hierarchical policies are shown to approach the optimal performance within a bounded error. 

Motivated by aforementioned ones, we designate a simple yet effective  sub-goal space that makes the hierarchy better interpretable. Meanwhile, to make the optimal policy expressible and learnable with the specified sub-goal space, we also leverage the benefits from the automatic temporal abstraction learning methods, which ultimately yields a hybrid system. 

\noindent{\bf Target-driven visual navigation.}  Deep reinforcement learning has been studied extensively for the target-driven visual navigation tasks \cite{ye2020seeing}.  These tasks can be categorized in terms of the description of the navigation target. \cite{zhu2017target, ye2018active} and \cite{kulhanek2019vision}  specify the navigation target by the image taken at the target location. The robotic object search task studied in \cite{mousavian2019visual,ye2019gaple,yang2018visual,druon2020visual} and the room navigation task introduced in \cite{wu2018building, wu2019bayesian} take the semantic label of the target object and room as the navigation target. The Embodied Question Answering \cite{das2018embodied, das2018neural, gordon2018iqa} and the Vision-and-Language Navigation \cite{anderson2018vision, wang2019reinforced} address the problem where the navigation target is provided with an unconstrained natural language. Here, we study the robotic object search task where the navigation target is an object specified by a semantic label.  
Unlike the previous work that plans over the atomic actions for navigation \cite{mousavian2019visual,ye2019gaple,yang2018visual,druon2020visual,wu2018building}, we learn a hierarchical policy that performs the robotic object search task in a more interpretable way. While \cite{das2018neural, gordon2018iqa} and \cite{wu2019bayesian} also study hierarchical policies, their low-level policies focus only on the sub-tasks without keeping the final navigation target in mind, thus may yield less optimal policies towards the final navigation target.

Notably, many of the previous works address the sparse reward issue by introducing additional supervision under the assumption that the robot can access the full information of the environments during the training time, such as defining the reward function with the distance between the robot's current location and the target location (a.k.a. reward shaping) \cite{wu2018building,mousavian2019visual}, adopting shortest path as the supervised signal for pre-training \cite{das2018embodied,wang2019reinforced}, and/or gradually increasing the distance between robot's starting location and the target location (a.k.a. curriculum learning) \cite{das2018neural,kulhanek2019vision}.  Nevertheless, for applications in real-world environments,  collecting all the information is unarguably expensive and sometimes impractical. {\it We would like to stress upon the point that our model does not assume any environment information available even during the training stage, which makes our object search task significantly more challenging.}

\section{Our Approach}
\label{sec:approach}
First, we define  the robotic object search task.
Formally speaking, when a target object is specified and provided with a semantic label, the robot is asked to search and approach the object from its random starting position. The RGB image from the robot's on-board camera is the only source of information for decision making. None of the environment information, such as the map of the environment or the location of the target object could be accessed. Once the area of the target object in the robot's viewpoint (the image captured by its camera) is larger than a predefined threshold, the robot stops and we consider it as a success. In this work, we present a novel two-layer hierarchical policy for the robot to perform the object search task, motivated by how human beings typically conduct object search. In the following sections, we first describe the hierarchy of policies. Then we introduce two kinds of reward functions, i.e. extrinsic rewards and intrinsic rewards, and we make use of these two reward functions to formulate the solution. Finally, we describe the network architecture adopted for learning the two-layer hierarchical policy.

\subsection{Hierarchy of Policies}
Our hierarchical policy has two levels, a high-level policy $\pi_h$ and a low-level policy $\pi_l$. At time step $t$, the robot takes the image captured by its camera as the current state $s_t$. Given a target object or goal $g$, the high-level layer proposes a sub-goal $sg_t \sim \pi_h(sg | s_t, g)$ and the low-level layer takes over the control. The low-level layer then draws an atomic action $a_t \sim \pi_l(a | s_t, g, sg_t)$ to perform. The robot will receive a new image/state $s_{t+1}$. The low-level layer repeats $N_t$ times till 1) the low-level layer terminates itself following the termination signal $term(s_{t+N_t}, g, sg_t)$; 2) the low-level layer achieves the sub-goal $sg_t$.
3) the low-level layer has performed a predefined maximum number of atomic actions.
Either way, the low-level layer terminates at state $s_{t+N_t}$, and then returns the control back to the high-level layer, and the high-level layer proposes another sub-goal. This process repeats until 1) the goal $g$ is achieved, i.e. the robot finds the target object successfully; 2) a predefined maximum number of atomic actions has been performed.

For the object search task, we define the sub-goal space as \emph{\{approach $obj |  obj$ is visible in the robot's current view\}}. We argue three reasons for the  sub-goal space definition, a) approaching an object that shows in the robot's view is a more general and relatively trainable task shown by \cite{ye2019gaple}. It also aligns well with the objective of the hierarchical reinforcement learning by breaking down the task into several easier sub-tasks; b) approaching a related object may increase the probability of seeing the target object. As soon as the target object is captured in the robot's current view,  the task becomes an object approaching task; c) as also suggested by \cite{kulkarni2016hierarchical}, specifying sub-goals over entities and relations can provide an efficient space for exploration in a complex environment. Moreover, in case there is no object visible in the robot's current view, we supplement a back-up ``random'' sub-goal invoking a random low-level policy. The atomic action space for the low-level layer is defined for navigation purpose, namely \emph{\{move forward / backward / left / right, turn left / right\}} in which the \emph{move} action updates the robot's location only and the \emph{turn} action drives the robot's rotation only.

\subsection{Extrinsic Rewards and Intrinsic Rewards}
\label{sec_rewards}
We define two kinds of reward functions. The extrinsic rewards $r^e$ are defined for our object search task, thus are goal dependent. Further, we also introduce the intrinsic rewards $r^i$ for the low-level sub-tasks. The intrinsic rewards are hereby sub-goal dependent. We specify the two reward functions respectively as follows. 

{\bf Extrinsic rewards $r^e$.} Without loss of generality, to encourage the robot to finish the object search task, we provide a positive extrinsic reward (in practice, $1$) when the robot reaches the final goal state. At all other intermediate states, the extrinsic rewards are set to $0$. Formally, $r^e_t(s_{t-1}, a_{t-1}, s_t, g) = 1 $ if and only if $s_t$ is a goal state of the goal $g$, otherwise $r^e_t(s_{t-1}, a_{t-1}, s_t, g) = 0 $.


{\bf Intrinsic rewards $r^i$.} To facilitate the robot perform the sub-task, i.e. approaching the object specified in the proposed sub-goal $sg$ which shows in the robot's current view, we adopt the similar binary rewards. To be specific, the intrinsic reward $r^i_t(s_{t-1}, a_{t-1}, s_t, sg) = 1$ if and only if $s_t$ is a goal state of the sub-goal $sg$, otherwise $r^i_t(s_{t-1}, a_{t-1}, s_t, sg) = 0$.

\subsection{Model Formulation}

\begin{figure*}[ht!]
\centering
\includegraphics[width=0.98\textwidth]{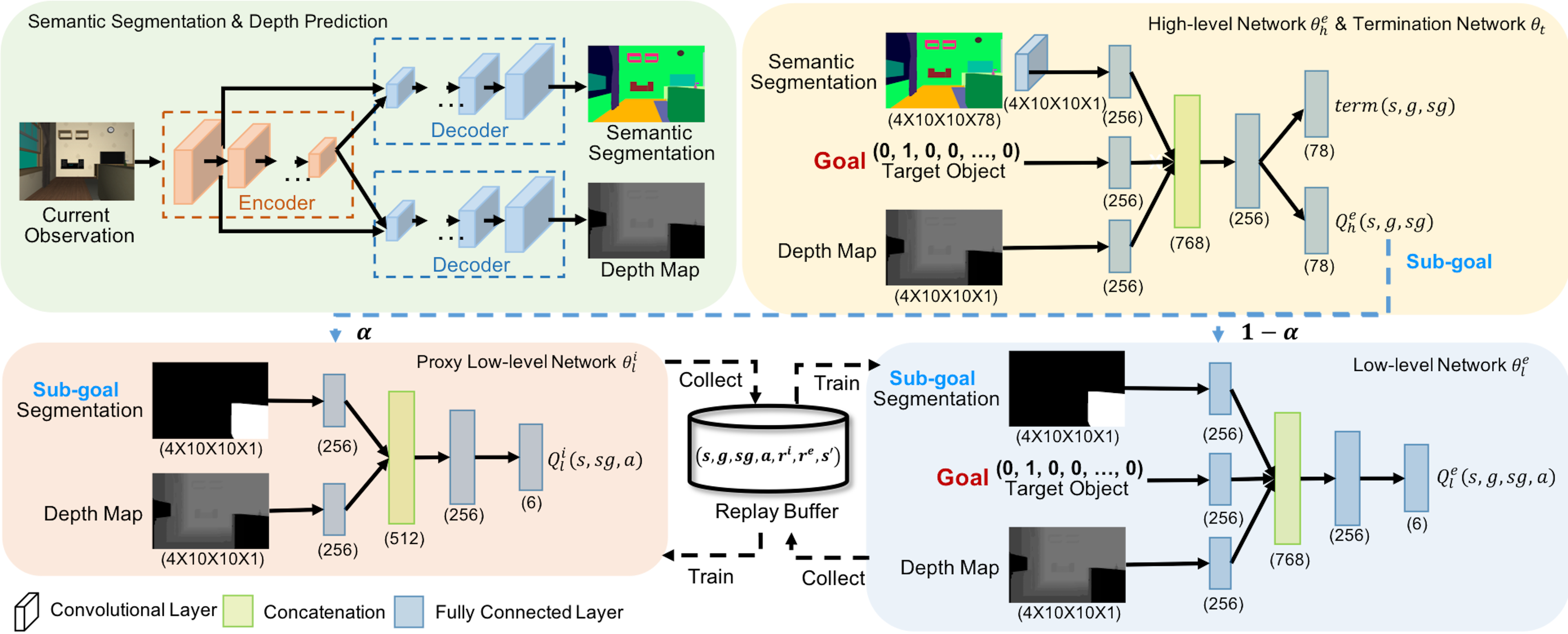}
\caption{Network architecture of our hierarchical reinforcement learning model.}
\label{fig:network}
\end{figure*}

We formulate the object search task in terms of the two rewards introduced in Sec.~\ref{sec_rewards}. When the robot starts from an initial state $s_0$, it proposes a sub-goal $sg_0$ aiming to achieve the final goal $g$ (locating and approaching the target object). To achieve the final goal, we can optimize the discounted cumulative extrinsic rewards, expected over all trajectories starting at state $s_0$ and sub-goal $sg_0$, which is $\mathbb{E} [\sum_{t=0}^{\infty}\gamma^tr^e_{t+1} |s_0, g, sg_0 ]$. If and only if the robot takes minimal steps to the goal state, the discounted cumulative extrinsic rewards are thus maximized.

The discounted cumulative extrinsic rewards is also known as the state action value $Q_h^e$ \cite{sutton2018reinforcement} for our high-level layer, i.e. $\mathbb{E} [\sum_{t=0}^{\infty}\gamma^tr^e_{t+1} | s_0=s, g=g, sg_0=sg] = Q_h^e(s,g, sg)$. Following the option-critic framework \cite{bacon2017option}, we unroll the $Q_h^e(s,g, sg)$ as,
\begin{equation}
\label{eq:qh_form}
\small
\begin{aligned}
&Q_h^e(s,g, sg) \\
= &\sum_a \pi_l(a| s, g, sg)\mathbb{E} [\sum_{t=0}^{\infty}\gamma^tr^e_{t+1} | s_0\!=\!s, g\!=\!g, sg_0\!=\!sg, a_0\!=\!a] \\
= &\sum_a \pi_l(a| s, g, sg)Q_l^e(s,g,sg,a),
\end{aligned}
\end{equation}
where the state action value $Q_l^e(s, g, sg, a)$ for our low-level layer is the discounted cumulative extrinsic rewards after taking action $a$ under the state $s$, goal $g$ and sub-goal $sg$. Given the transition probability $P(s'| s,a)$ which denotes the probability of being state $s'$ after taking action $a$ at state $s$,  $Q_l^e(s, g, sg, a)$ can be further formulated as,
\begin{equation}
\small
Q_l^e(s,g,sg,a) = \sum_{s'}P(s'| s,a)[r^e(s,a,s',g) + \gamma  U(g,sg,s')], \nonumber
\end{equation}
\begin{equation}
\small
\begin{aligned}
U(g,sg,s') = (1 - &term(s',g,sg)) Q_h^e(s',g,sg) + \\&term(s',g,sg) V_h^e(s',g), \nonumber 
\end{aligned}
\end{equation}
\begin{equation}
\small
V_h^e(s',g) = \sum_{sg'} \pi_h(sg'| s',g)Q_h^e(s',g,sg').
\end{equation}





We parameterize $Q^e_h(s,g,sg)$, $Q^e_l(s,g,sg,a)$ and $term(s,g,sg)$ with $\theta^e_h$, $\theta^e_l$ and $\theta_t$ respectively. Then the high-level policy $\pi_h(sg|s,g)=\mathds{1}(sg=\argmax_{sg}Q^e_h(s,g,sg))$, and $\pi_l(a|s,g,sg)=\mathds{1}(a=\argmax_{a}Q^e_l(s,g,sg,a))$ is our low-level policy. We adopt the DQN \cite{mnih2015human} based method to learn $Q^e_h(s,g,sg)$ and $Q^e_l(s,g,sg,a)$ in which we update both of the values towards the $1$-step extrinsic return $R_1^e = r^e(s, a, s', g)+\gamma U(g, sg, s')$, and consequently $\theta^e_h$ and $\theta^e_l$ can be updated by Equation~\ref{eq:he_update} and \ref{eq:le_update}. In addition, $\theta_t$ can be updated by Equation~\ref{eq:t_update} as demonstrated by \cite{bacon2017option}.

\begin{equation}
\label{eq:he_update}
\small
\theta^e_h \gets \theta^e_h - \nabla_{\theta^e_h}[R_1^e-Q_{\theta^e_h}(s,g,sg)]^2.
\end{equation}
\begin{equation}
\label{eq:le_update}
\small
\theta^e_l \gets \theta^e_l - \nabla_{\theta^e_l}[R_1^e-Q_{\theta^e_l}(s,g,sg,a)]^2.
\end{equation}
\begin{equation}
\small
\label{eq:t_update}
\theta_t \gets \theta_t - \nabla_{\theta_t}term_{\theta_t}(s',g,sg)(Q^e_h(s',g, sg)-V^e_h(s',g)).
\end{equation}


Since the robot may start at a position far away from the target object, it is unlikely for the robot to encounter the sparse extrinsic rewarding states through the $\epsilon$-greedy \cite{mnih2015human} exploration policy and collect the experience samples to effectively train $\theta^e_h$, $\theta^e_l$ and $\theta_t$. On the contrary, encountering the intrinsic rewarding states is much more possibly as an object shows in the robot's current view is usually nearby. Therefore, training the robot to achieve a sub-goal is more accessible. Then, by iteratively asking the robot to achieve suitable sub-goals, i.e. to approach related objects, the robot is more likely to observe the target object and collect the valuable experience samples to train $\theta^e_h$, $\theta^e_l$ and $\theta_t$.

We hereby define a proxy low-level policy $\pi^p_l(a|s,sg)$ aiming to achieve the proposed sub-goal $sg$. Similarly, we learn the proxy low-level policy by optimizing the discounted cumulative intrinsic rewards $Q^i_l(s,sg,a)$. 
We adopt the DQN method \cite{mnih2015human} to learn it by updating its parameter $\theta^i_l$ with Equation~\ref{eq:li_update}, where $R^i_1 = r^i(s, a, s', sg) + \gamma \max_{a}Q^i_l(s',sg, a)$ is the $1$-step intrinsic return. As a result, $\pi^p_l(a|s,sg) =\mathds{1}(a=\argmax_{a}Q^i_l(s,sg,a))$.


\begin{equation}
\label{eq:li_update}
\small
\theta^i_l \gets \theta^i_l - \nabla_{\theta^i_l}[R_1^i-Q_{\theta^i_l}(s,sg,a)]^2.
\end{equation}

For our low-level layer to balance between exploitation by achieving the goal $g$ with the policy $\pi_l(a|s,g,sg)$ and the exploration by achieving the sub-goal $sg$ with the proxy policy $\pi^p_l(a|s,sg)$, we introduce a hyper-parameter $\alpha \in [0,1]$ as the probability that the low-level layer adopts the proxy policy $\pi^p_l(a|s,sg)$ to explore the environment and collect the experience samples. The experience samples are used to batch train $\theta^e_h$, $\theta^e_l$, $\theta_t$ and $\theta^i_l$ with Equation~\ref{eq:he_update}, \ref{eq:le_update}, \ref{eq:t_update} and \ref{eq:li_update} respectively.  In practice, $\alpha$ decays from $1$ to $0$ across the training episodes to enable our low-level layer to act optimally towards the goal with the policy $\pi_l(a|s,g,sg)$ eventually.

\subsection{HIEM Network Architecture}
\label{sec:network}


Since the image captured by the robot's on-board camera serves as the robot's current state, we adopt deep neural networks as $\theta^e_h$, $\theta^e_l$, $\theta_t$ and $\theta^i_l$ to handle the high dimensional inputs and approximate $Q^e_h(s,g,sg)$, $Q^e_l(s,g,sg,a)$, $term(s,g,sg)$ and $Q^i_l(s,sg,a)$.

Fig.~\ref{fig:network} illustrates our network architecture. 
For the object search task, semantic segmentation and depth map are necessary for the robot to detect the target object and avoid collision during the navigation. Therefore, we first adopt the encoder-decoder network \cite{ye2019gaple} to predict the semantic segmentation and the depth map from the robot's observation. We take the predicted results as the inputs to our policy networks to avoid the need of visual domain adaption \cite{mousavian2019visual}. The predicted results of the $4$ history observations are fed into our high-level network $\theta^e_h$ in addition to a one-hot vector representing the target object. The channel size of the segmentation input is first reduced to $1$ through a convolutional layer with $1$ filter of kernel size $1\times1$, and then the three inputs are fed into three different fully connected layers respectively and their outputs are further concatenated into a joint vector before attaching another fully connected layer to generate an embedding fusion. Our high-level network $\theta^e_h$ feeds the embedding fusion into one additional fully connected layer to approximate $Q^e_h(s,g,sg)$. To save the number of parameters, our termination network $\theta_t$ shares most parameters with the high-level network $\theta^e_h$ except the last fully connected layer where it adopts a new one to approximate $term(s,g,sg)$.

For the low-level network $\theta^e_l$ and $\theta^i_l$, we take the sub-goal specified channel of the predicted semantic segmentation and the predicted depth map as the inputs. The low-level network $\theta^e_l$ takes the one-hot vector of the target object as an additional input. Similar to our high-level network, each input of $\theta^e_l$ and $\theta^i_l$ is fed into a fully connected layer before being concatenated together to generate an embedding fusion with a new fully connected layer. The embedding fusion is further fed into an additional fully connected layer to approximate $Q^e_l(s,g,sg,a)$ and $Q^i_l(s,sg,a)$.

We follow Equation~\ref{eq:he_update}, \ref{eq:le_update}, \ref{eq:t_update} and \ref{eq:li_update} to learn $Q^e_h(s,g,sg)$, $Q^e_l(s,g,sg,a)$, $term(s,g,sg)$ and $Q^i_l(s,sg,a)$ respectively.

\section{Experiments}

\begin{figure*}[ht!]
\centering
\begin{tabular}{ccc}
\includegraphics[width=0.46\textwidth]{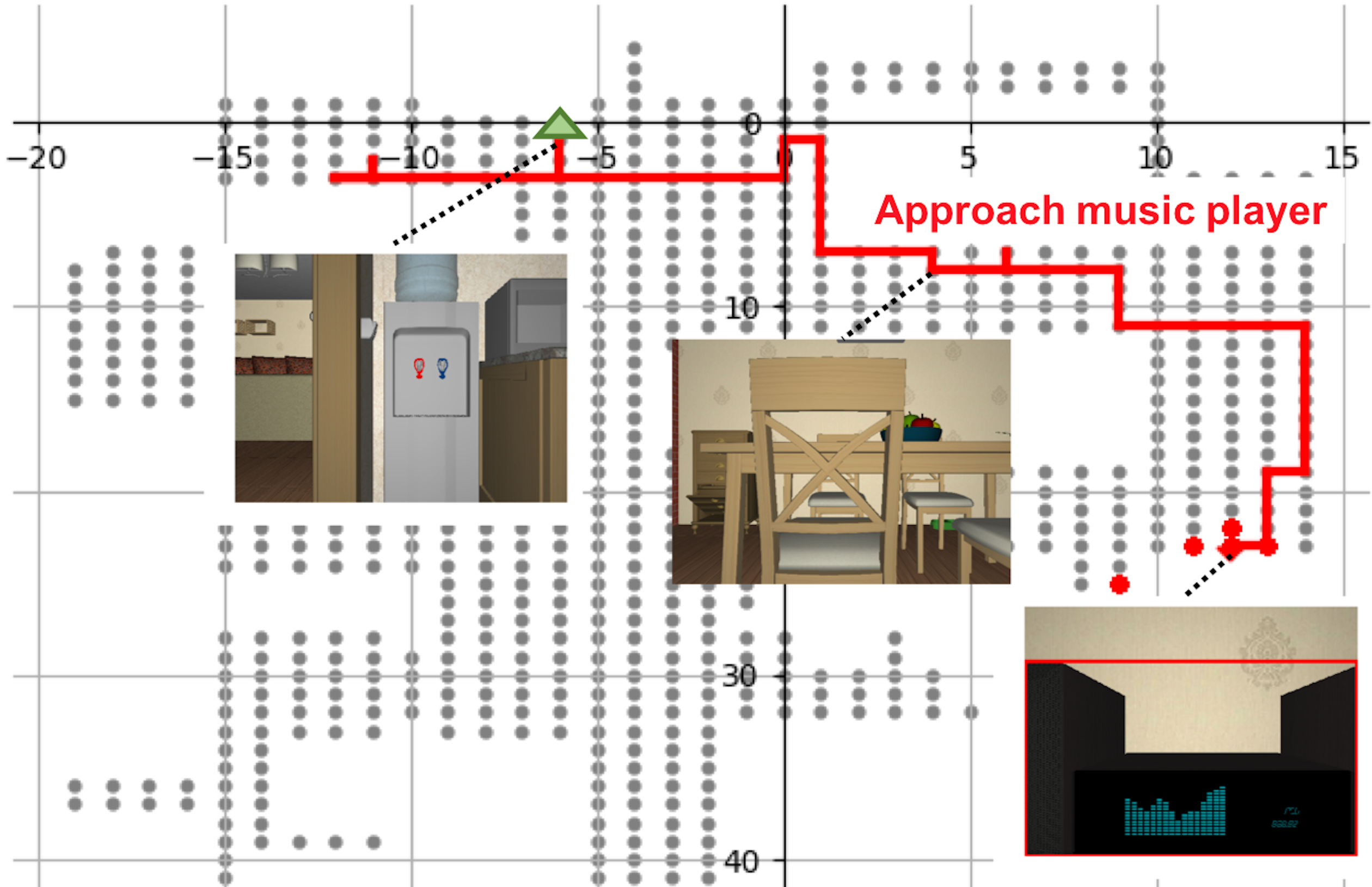} &&
\includegraphics[width=0.46\textwidth]{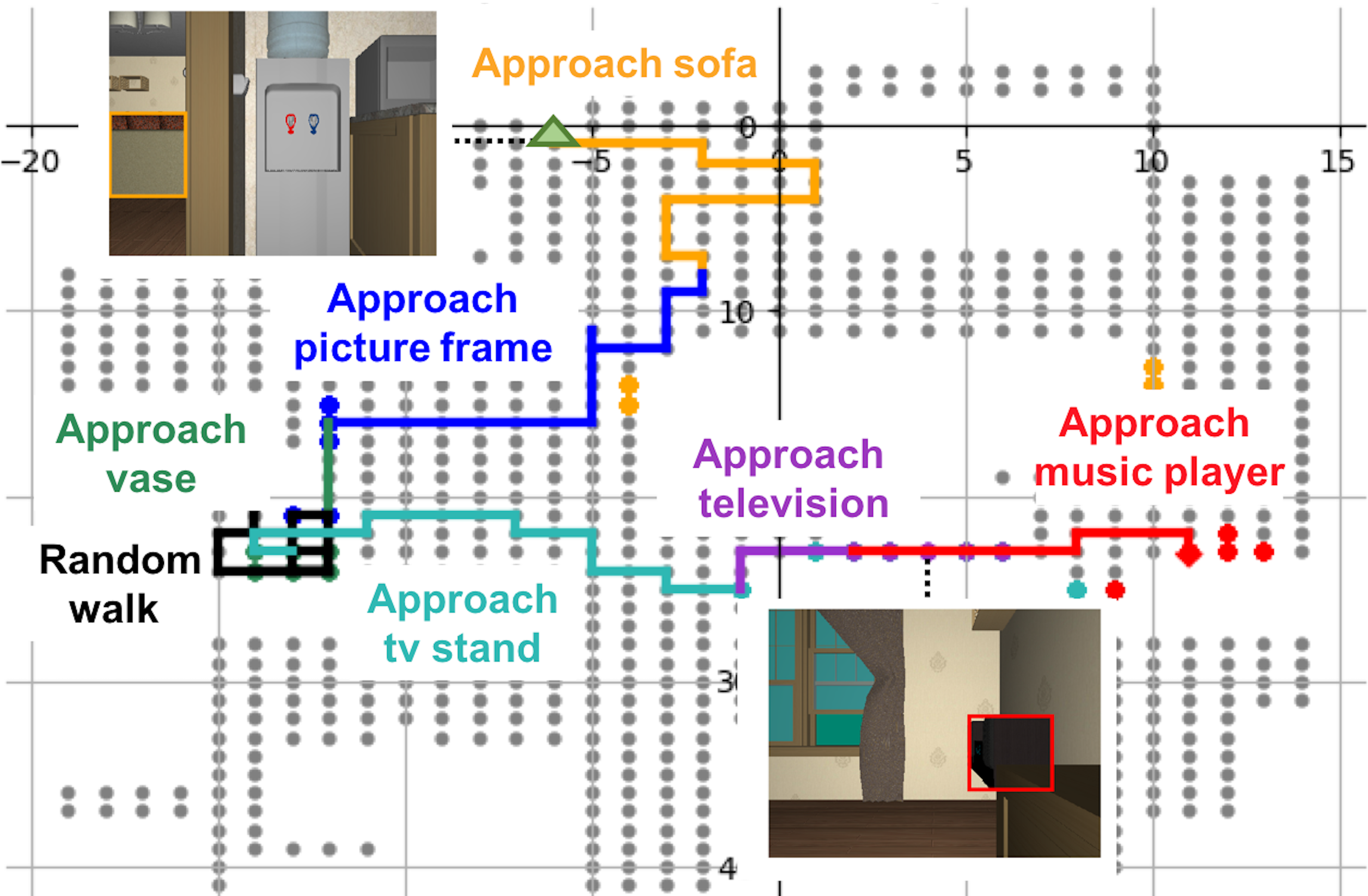} \\
\small{\textsc{DQN} ($203$ steps)}
&&\small{\textsc{h-DQN} ($146$ steps)} \\
\\
\includegraphics[width=0.46\textwidth]{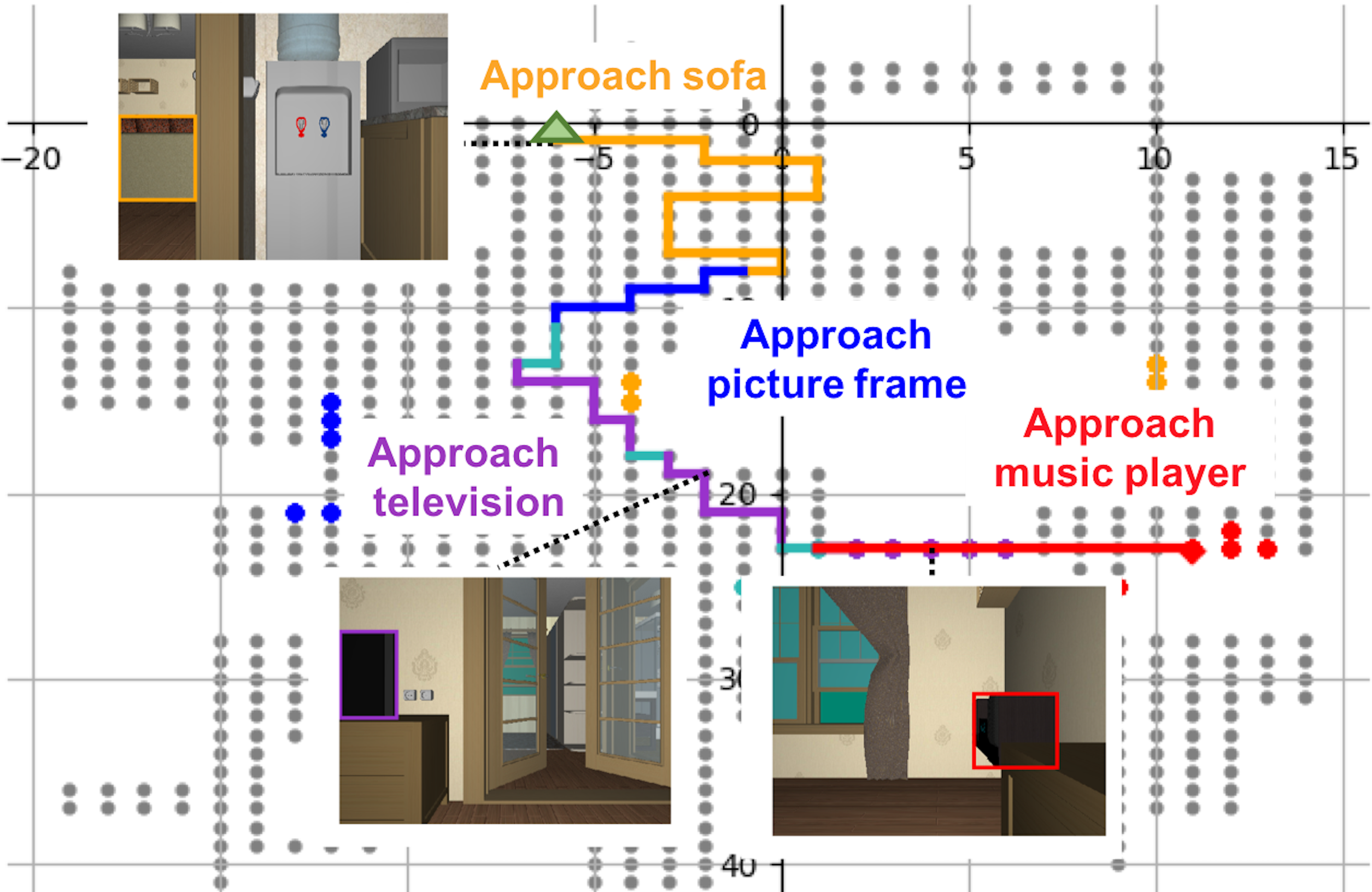} &&
\includegraphics[width=0.46\textwidth]{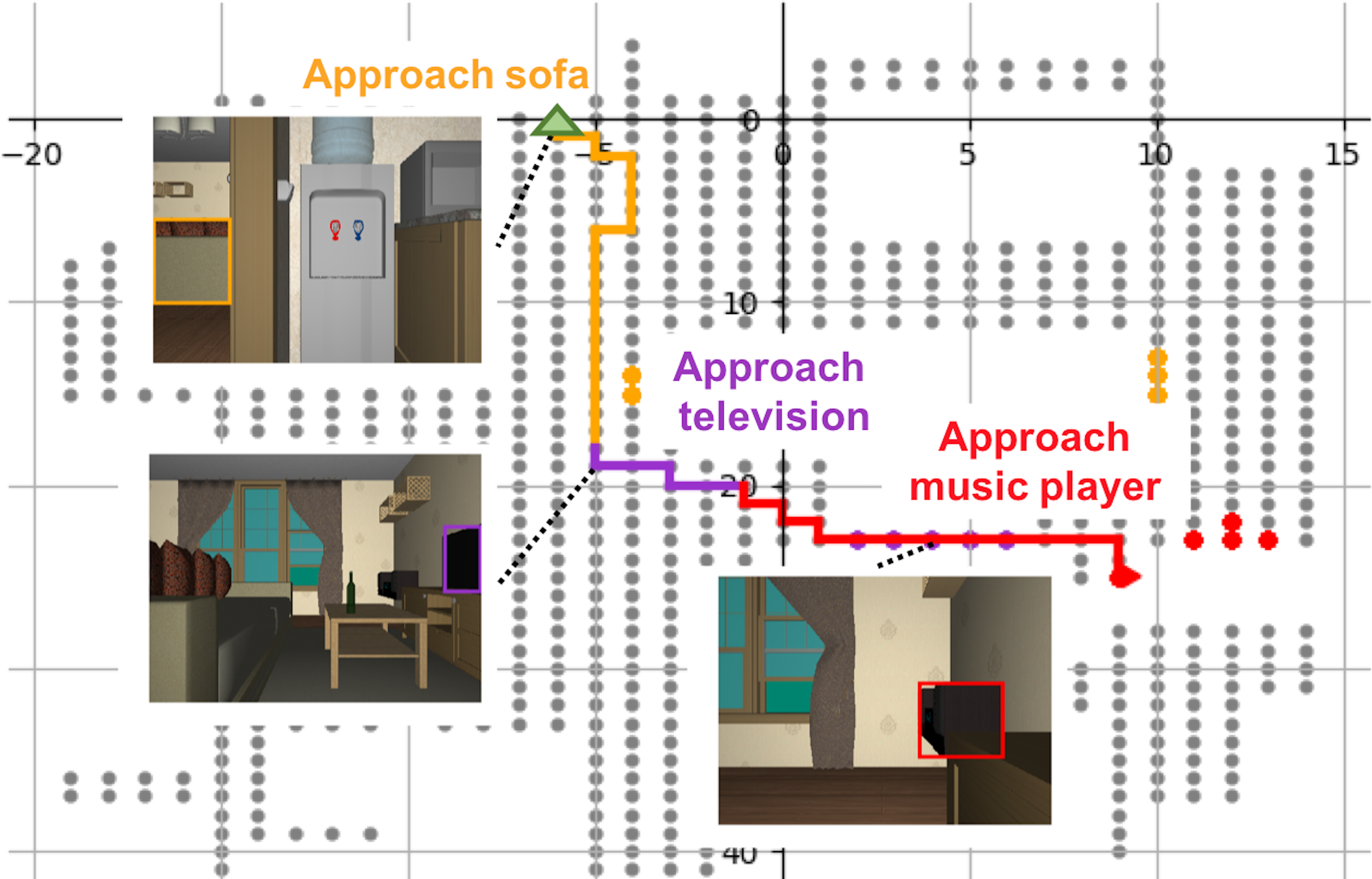} \\
\small{\textsc{HIEM}-low ($81$ steps)}
&&\small{\textsc{HIEM} ($44$ steps)}\\
\end{tabular}
\caption{Trajectories generated by \textsc{DQN} \cite{mnih2015human}, \textsc{h-DQN} \cite{kulkarni2016hierarchical} and our method \textsc{HIEM}-low and \textsc{HIEM} for searching the target object \emph{music player} (red dots) from the same starting position (green triangle) which is $39$ steps away. Different colors represent different sub-goals in which the colored lines and dots denote the corresponding sub-goal-oriented trajectories and sub-goal states respectively. Our method \textsc{HIEM} generates a more concise and interpretable trajectory. We refer readers to the supplemental video demo from \url{https://youtu.be/rAHB3jIS3Wo} for animated demonstrations.}
\label{fig:qua_results}
\end{figure*}

\subsection{Dataset}
We validate our framework on the 
simulation platform House3D \cite{wu2018building}. House3D consists of rich indoor environments with diverse layouts for a virtual robot to navigate. In each indoor environment, a variety of objects are scattered at many locations, such as \emph{television, sofa, desk}. While navigating, the robot has a first-person view RGB image as its observation. The simulator also provides the robot with the ground truth semantic segmentation and depth map corresponding to the RGB image.  The RGB images, as well as the semantic segmentation and depth maps can be used as the training data to learn the encoder-decoder network \cite{ye2019gaple} (shown in Fig.~\ref{fig:network} upper left) for semantic segmentation and depth prediction as we mentioned in Sec.~\ref{sec:network}. We refer interested readers to \cite{ye2019gaple} for more details.
In addition, the trained model, specifically the semantic segmentation prediction, can be used as the robot's detection system.

To validate our proposed method in learning hierarchical policy for object search, we conduct the experiments in an indoor environment where the objects' placements are in accordance with the real-world scenario. For example, the \emph{television} is placed close to the \emph{sofa}, and is likely occluded by the \emph{sofa} at many viewpoints. In such a way, to search the target object \emph{television}, the robot could approach \emph{sofa} first to increase the likelihood of seeing the \emph{television}.

We consider discrete actions for the robot to navigate in this environment. Specifically, the robot moves forward / backward / left / right $0.2$ meters, or rotates $90$ degrees every time. We also discretize the environment into a certain number of reachable locations, as shown in Fig.~\ref{fig:qua_results}.

\subsection{Experimental Setting}

We compare the following methods and variants:

{\bf \textsc{Oracle}} and {\bf \textsc{Random}}. At each time step, the robot ignores its observation and performs the optimal action and a random action respectively. 

{\bf \textsc{A3C}} \cite{mnih2016asynchronous}. The vanilla A3C implementation that has been wildly adopted for the navigation task in the previous work \cite{zhu2017target,ye2018active,ye2019gaple,yang2018visual,druon2020visual}. It learns the action policy $\pi(a|s,g)$ and the state value $V^e(s,g)$ with a similar network architecture as our high-level network $\theta^e_h$.

{\bf \textsc{DQN}} \cite{mnih2015human}. The vanilla DQN implementation that adopts a similar network architecture as our high-level network $\theta^e_h$ to predict the state action value $Q^e(s, g, a)$.


{\bf \textsc{OC}} \cite{bacon2017option}. 
The Option-Critic implementation that learns a hierarchical policy autonomously by maximizing the discounted cumulative extrinsic rewards where only the number of the options needs to be manually set. We set it as $4$ in our experiments.

{\bf \textsc{h-DQN}} \cite{kulkarni2016hierarchical} with our proposed sub-goal space. It is equivalent to our method when we set $term(s,g,sg)=0$ and $\alpha=1$ to disable both the termination network $\theta_t$ and the low-level network $\theta^e_l$ .


{\bf \textsc{HIEM}}. Our method follows Sec~\ref{sec:approach}. To further identify the role of each component of our method, we conduct ablation studies by disabling one component at a time. Specifically, {\bf \textsc{HIEM}-proxy} sets $\alpha=0$ to disable the proxy low-level network  $\theta^i_l$, {\bf \textsc{HIEM}-low} sets $\alpha=1$ to disable the low-level network  $\theta^e_l$, and {\bf \textsc{HIEM}-term} sets $term(s,g,sg)=0$ to disable the termination network $\theta_t$.




For fair comparisons, all the methods share similar network architectures and hyperparameters, and they all take the predicted semantic segmentation and the depth map as the inputs. To be specific, for DQN networks in the method \textsc{DQN}, \textsc{h-DQN} and \textsc{HIEM}, we adopt the Double DQN \cite{van2015deep} technique where we train the main network every $100$ time steps with a batch of size $64$ and we update the target network every $100,000$ time steps. The exploration rate decreases from $1$ to $0.1$ over $10,000$ time steps. For the A3C network, we set the weight of the entropy regularization term as $0.01$ and we update the network for every $10$ time steps unrolled.
We adopt RMSProp optimizer of learning rate $1 \times 10^{-4}$ to train each method to search $6$ different target objects ($78$ in total) from random starting positions in the environment. 
During testing time, we randomly sample $100$ starting positions and the corresponding target objects. We set the maximum number of atomic actions that all methods can take as $500$, and for the method \textsc{h-DQN} and \textsc{HIEM}, the maximum number of atomic actions that the low-level layer can take at each time is $25$. The robot stops either when it reaches the goal state (success case) or when it runs out of $500$ atomic action steps (failure case). We implement all the methods using Tensorflow toolbox and conduct all the experiments with Nvidia V100 GPUs and $16$ Intel Xeon E5-2680 v4 CPU cores. In general, each training takes around $2$ days.

\subsection{Experimental Results and Discussion}

\begin{table}
\normalsize
\caption{The performance of all methods for the object search task. (SR: Success Rate; AS / MS: Average Steps / Minimal Steps over all successful cases; SPL: Success weighted by inverse Path Length; AR: Average discounted cumulative extrinsic Rewards.)}
\label{tbl:metric}
\begin{center}
\begin{tabular}{lcccc}
\specialrule{0.12em}{0pt}{1.5pt}
\specialrule{0em}{1pt}{1pt}
Method &SR$\uparrow$ & AS / MS$\downarrow$ & SPL$\uparrow$ & AR$\uparrow$ \\
\specialrule{0.12em}{1pt}{1.5pt}
\textsc{Oracle} &1.00 & 25.63 / 25.63 & 1.00 & 0.79\\
\specialrule{0.04em}{1pt}{1pt}
\textsc{Random} &0.19 & 188.11 /  7.05 & 0.03 &0.08 \\
\textsc{A3C} &0.13  &  93.23 /  4.00 &0.03 &0.08   \\
\textsc{DQN} &0.47 & 120.74 / 16.09 &0.20 &0.26  \\
\textsc{OC} &0.14 & 99.29 /  5.14 &0.06 &0.09  \\
\textsc{h-DQN} &0.74 &182.15 / 23.62 &0.17 &0.23 \\
\specialrule{0.04em}{1pt}{1pt}
\textbf{Ours} &&&& \\
\quad \textsc{HIEM}-proxy &0.40 &95.08 / 15.03 &0.12 &0.22\\
\quad \textsc{HIEM}-low &0.99 & 76.81 / 25.55 & 0.47 &0.56\\
\quad \textsc{HIEM}-term & \textbf{1.00} &49.42 / 25.63 &0.65 &0.66\\
\quad \textsc{HIEM} &\textbf{1.00} & \textbf{41.18 / 25.63} & \textbf{0.72} &\textbf{0.70}\\
\specialrule{0.12em}{1.5pt}{0pt}
\end{tabular}
\end{center}
\end{table}

Since we formulate the object search problem as maximizing the discounted cumulative extrinsic rewards, we take the Average discounted cumulative extrinsic Rewards (AR) as one of the evaluation metrics, calculated by:
\begin{equation}
\small
\frac{1}{N}\sum_{i=1}^{N}\sum_{t=0}^{\infty}\gamma^t r^e_{t+1} = \frac{1}{N}\sum_{i=1}^{N}\mathds{1}(success)\gamma^{\#steps}*1,
\end{equation}
where $\gamma \in (0,1]$ is the discount factor. From the perspective of the evaluation metric, it can also be seen as a trade-off between the success rate metric and the average steps metric. With the higher value of $\gamma$, the average steps metric weighs less, and vice versa. In our experiments, we set $\gamma = 0.99$.

In addition, we also report the following widely used evaluation metrics. Success Rate (SR). Average Steps over all successful cases compared to the Minimal Steps over these cases (AS / MS). Success weighted by inverse Path Length (SPL) \cite{anderson2018evaluation}, which is calculated as $\frac{1}{N}\sum_{i=1}^{N}S_i \frac{l_i}{max(l_i, p_i)}$. Here, $S_i$ is the binary indicator of success in episode $i$, $l_i$ and $p_i$ are the lengths of the shortest path and the path actually taken by the robot. We adopt the number of the action steps as the path length. As a result, SPL also trades-off success rate against average steps.

Table~\ref{tbl:metric} shows comparisons of all the methods in performing the object search task. It demonstrates the superiority of our method over all metrics, and also highlights the following observations.

{\bf The intrinsic rewards help to explore.} Comparing to \textsc{h-DQN} and our methods (\textsc{HIEM}, \textsc{HIEM}-low, \textsc{HIEM}-term) which model the object search task with both extrinsic and intrinsic rewards, all the other methods where no intrinsic rewards is involved achieve unsatisfactory success rate. It indicates that under the sparse extrinsic rewards setting, the robot struggles to reach the goal state even with the hierarchical policy \textsc{OC} or \textsc{HIEM}-proxy, while our intrinsic rewards effectively encourage the robot to explore the environment and encounter the goal state. In fact, the intrinsic rewards guide our proxy low-level network to approach a visible object, and only after the proxy low-level network achieves good performance can it collaborate with our high-level network to help explore. 

{\bf Our intrinsic-extrinsic modeling contributes to a more optimal policy.} Though our intrinsic rewards help to explore the environment and improve the success rate, they are limited in improving the policy in terms of the optimality, as suggested by the higher AS and lower SPL and AR that \textsc{h-DQN} and \textsc{HIEM}-low achieve in comparison with \textsc{HIEM}. Different from \textsc{h-DQN} or \textsc{HIEM}-low that models the low-level layer with the intrinsic rewards solely, our \textsc{HIEM} adopts the novel intrinsic-extrinsic modeling and yields a more optimal policy,  demonstrating the role of our intrinsic-extrinsic modeling in learning an optimal policy.

{\bf Early termination to the non-optimal low-level policy is necessary. } A non-optimal low-level policy would drive the robot to an undesirable state that in consequence hurts the object search performance. The issue is shown to be mitigated by terminating the low-level policy at a valuable state in \textsc{HIEM}-low and  \textsc{HIEM} when comparing them with \textsc{h-DQN} and \textsc{HIEM}-term respectively. Furthermore, we also observe that the termination function helps more to less optimal low-level policy as more improvements are achieved from \textsc{h-DQN} to \textsc{HIEM}-low.

\begin{table}[ht]
   \caption{Average SPL achieved by all methods on $4$ random environments.}
   \label{tbl:avg_spl}
    \centering
    \begin{tabular}{lcccccc}
    \specialrule{0.1em}{0pt}{1.5pt}
         Method & \textsc{Random} & \textsc{A3C} & \textsc{DQN} & \textsc{OC} & \textsc{h-DQN} & \textbf{\textsc{HIEM}}  \\
         \specialrule{0.04em}{1pt}{1pt}
         Avg SPL &0.03 &0.03 &0.35 &0.03 &0.11 &\textbf{0.54}\\
    \specialrule{0.1em}{0pt}{1.5pt}
    \end{tabular}
\end{table}
We also report in Table~\ref{tbl:avg_spl} the average SPL achieved by all methods on $4$ random environments. It further validates the superiority of our \textsc{HIEM} on other environments as well.
We depict sample qualitative results in Fig.~\ref{fig:qua_results}, which shows that our method yields a more concise and interpretable trajectory compare to other methods for the object search task.


\section{Conclusion and Future Work}

In this paper, we present a novel two-layer hierarchical policy learning framework for the robotic object search task. The hierarchical policy builds on a simple yet effective and interpretable low dimensional sub-goal space, and is learned with both extrinsic  and intrinsic rewards to perform the object search task in a more optimal and interpretable way.
When our high-level layer plans over the specified sub-goal space, the low-level layer plans over the atomic actions to collaborate with the high-level layer to better achieve the goal. This is efficiently learned with the experience samples collected by our proxy low-level policy, a policy optimizes towards the proposed sub-goals. Moreover, our low-level layer terminates at valuable states which further approximates the optimal policy. The empirical and extensive experiments together with the ablation studies on House3D platform demonstrate the efficacy and efficiency of our presented framework. The presented HIEM framework further paves several possible avenues for future study. A promising one is by incorporating the Goals Relational Graph (GRG) \cite{ye2021GRG} to integrate top-down human knowledge together with the human specified sub-goal space to facilitate the object search with improved efficiency. 


We want to mention that the current work assumes the robot can access the environment for training before being deployed in the same one for object search. In other words, we do not aim for the generalization ability towards novel environments, but our success sheds light on how to generalize well. Specifically, an optimal object search policy in an environment is determined by the map of the environment. In order to generalize a learned object search policy to a new environment where the map is unknown and no extra exploration or training process is allowed, the robot must be able to infer the map from its observation and/or from its external memory or knowledge. While the large high-resolution map is extremely challenging to infer, inferring a small part of it and a low-resolution object arrangement are still tractable, which in consequence makes both of our low-level policy and high-level policy more likely to generalize well. We deem it as our future work.


\bibliography{references}

\begin{thebibliography}{10}
\providecommand{\url}[1]{#1}
\csname url@samestyle\endcsname
\providecommand{\newblock}{\relax}
\providecommand{\bibinfo}[2]{#2}
\providecommand{\BIBentrySTDinterwordspacing}{\spaceskip=0pt\relax}
\providecommand{\BIBentryALTinterwordstretchfactor}{4}
\providecommand{\BIBentryALTinterwordspacing}{\spaceskip=\fontdimen2\font plus
\BIBentryALTinterwordstretchfactor\fontdimen3\font minus
  \fontdimen4\font\relax}
\providecommand{\BIBforeignlanguage}[2]{{%
\expandafter\ifx\csname l@#1\endcsname\relax
\typeout{** WARNING: IEEEtran.bst: No hyphenation pattern has been}%
\typeout{** loaded for the language `#1'. Using the pattern for}%
\typeout{** the default language instead.}%
\else
\language=\csname l@#1\endcsname
\fi
#2}}
\providecommand{\BIBdecl}{\relax}
\BIBdecl

\bibitem{das2018embodied}
A.~Das, S.~Datta, G.~Gkioxari, S.~Lee, D.~Parikh, and D.~Batra, ``Embodied
  question answering,'' in \emph{Proceedings of the IEEE Conference on Computer
  Vision and Pattern Recognition Workshops}, 2018, pp. 2054--2063.

\bibitem{kojima2019learn}
N.~Kojima and J.~Deng, ``To learn or not to learn: Analyzing the role of
  learning for navigation in virtual environments,'' \emph{arXiv preprint
  arXiv:1907.11770}, 2019.

\bibitem{mishkin2019benchmarking}
D.~Mishkin, A.~Dosovitskiy, and V.~Koltun, ``Benchmarking classic and learned
  navigation in complex 3d environments,'' \emph{arXiv preprint
  arXiv:1901.10915}, 2019.

\bibitem{arulkumaran2017brief}
K.~Arulkumaran, M.~P. Deisenroth, M.~Brundage, and A.~A. Bharath, ``A brief
  survey of deep reinforcement learning,'' \emph{arXiv preprint
  arXiv:1708.05866}, 2017.

\bibitem{mirowski2016learning}
P.~Mirowski, R.~Pascanu, F.~Viola, H.~Soyer, A.~J. Ballard, A.~Banino,
  M.~Denil, R.~Goroshin, L.~Sifre, K.~Kavukcuoglu \emph{et~al.}, ``Learning to
  navigate in complex environments,'' \emph{arXiv:1611.03673}, 2016.

\bibitem{zhu2017target}
Y.~Zhu, R.~Mottaghi, E.~Kolve, J.~J. Lim, A.~Gupta, L.~Fei-Fei, and A.~Farhadi,
  ``Target-driven visual navigation in indoor scenes using deep reinforcement
  learning,'' in \emph{2017 IEEE international conference on robotics and
  automation (ICRA)}.\hskip 1em plus 0.5em minus 0.4em\relax IEEE, 2017, pp.
  3357--3364.

\bibitem{gu2017deep}
S.~Gu, E.~Holly, T.~Lillicrap, and S.~Levine, ``Deep reinforcement learning for
  robotic manipulation with asynchronous off-policy updates,'' in \emph{2017
  IEEE International Conference on Robotics and Automation (ICRA)}.\hskip 1em
  plus 0.5em minus 0.4em\relax IEEE, 2017, pp. 3389--3396.

\bibitem{popov2017data}
I.~Popov, N.~Heess, T.~Lillicrap, R.~Hafner, G.~Barth-Maron, M.~Vecerik,
  T.~Lampe, Y.~Tassa, T.~Erez, and M.~Riedmiller, ``Data-efficient deep
  reinforcement learning for dexterous manipulation,'' \emph{arXiv preprint
  arXiv:1704.03073}, 2017.

\bibitem{rajeswaran2017learning}
A.~Rajeswaran, V.~Kumar, A.~Gupta, G.~Vezzani, J.~Schulman, E.~Todorov, and
  S.~Levine, ``Learning complex dexterous manipulation with deep reinforcement
  learning and demonstrations,'' \emph{arXiv preprint arXiv:1709.10087}, 2017.

\bibitem{chen2017socially}
Y.~F. Chen, M.~Everett, M.~Liu, and J.~P. How, ``Socially aware motion planning
  with deep reinforcement learning,'' in \emph{2017 IEEE/RSJ International
  Conference on Intelligent Robots and Systems (IROS)}.\hskip 1em plus 0.5em
  minus 0.4em\relax IEEE, 2017, pp. 1343--1350.

\bibitem{everett2018motion}
M.~Everett, Y.~F. Chen, and J.~P. How, ``Motion planning among dynamic,
  decision-making agents with deep reinforcement learning,'' in \emph{2018
  IEEE/RSJ International Conference on Intelligent Robots and Systems
  (IROS)}.\hskip 1em plus 0.5em minus 0.4em\relax IEEE, 2018, pp. 3052--3059.

\bibitem{mnih2015human}
V.~Mnih, K.~Kavukcuoglu, D.~Silver, A.~A. Rusu, J.~Veness, M.~G. Bellemare,
  A.~Graves, M.~Riedmiller, A.~K. Fidjeland, G.~Ostrovski \emph{et~al.},
  ``Human-level control through deep reinforcement learning,'' \emph{Nature},
  vol. 518, no. 7540, p. 529, 2015.

\bibitem{abbeel2004apprenticeship}
P.~Abbeel and A.~Y. Ng, ``Apprenticeship learning via inverse reinforcement
  learning,'' in \emph{Proceedings of the twenty-first international conference
  on Machine learning}.\hskip 1em plus 0.5em minus 0.4em\relax ACM, 2004, p.~1.

\bibitem{mousavian2019visual}
A.~Mousavian, A.~Toshev, M.~Fi{\v{s}}er, J.~Ko{\v{s}}eck{\'a}, A.~Wahid, and
  J.~Davidson, ``Visual representations for semantic target driven
  navigation,'' in \emph{2019 International Conference on Robotics and
  Automation (ICRA)}.\hskip 1em plus 0.5em minus 0.4em\relax IEEE, 2019, pp.
  8846--8852.

\bibitem{wang2018look}
X.~Wang, W.~Xiong, H.~Wang, and W.~Yang~Wang, ``Look before you leap: Bridging
  model-free and model-based reinforcement learning for planned-ahead
  vision-and-language navigation,'' in \emph{Proceedings of the European
  Conference on Computer Vision (ECCV)}, 2018, pp. 37--53.

\bibitem{wang2019reinforced}
X.~Wang, Q.~Huang, A.~Celikyilmaz, J.~Gao, D.~Shen, Y.-F. Wang, W.~Y. Wang, and
  L.~Zhang, ``Reinforced cross-modal matching and self-supervised imitation
  learning for vision-language navigation,'' in \emph{Proceedings of the IEEE
  Conference on Computer Vision and Pattern Recognition}, 2019, pp. 6629--6638.

\bibitem{ye2018active}
X.~Ye, Z.~Lin, H.~Li, S.~Zheng, and Y.~Yang, ``Active object perceiver:
  Recognition-guided policy learning for object searching on mobile robots,''
  in \emph{2018 IEEE/RSJ International Conference on Intelligent Robots and
  Systems (IROS)}.\hskip 1em plus 0.5em minus 0.4em\relax IEEE, 2018, pp.
  6857--6863.

\bibitem{nachum2019does}
O.~Nachum, H.~Tang, X.~Lu, S.~Gu, H.~Lee, and S.~Levine, ``Why does hierarchy
  (sometimes) work so well in reinforcement learning?'' \emph{arXiv preprint
  arXiv:1909.10618}, 2019.

\bibitem{kulkarni2016hierarchical}
T.~D. Kulkarni, K.~Narasimhan, A.~Saeedi, and J.~Tenenbaum, ``Hierarchical deep
  reinforcement learning: Integrating temporal abstraction and intrinsic
  motivation,'' in \emph{Advances in neural information processing systems},
  2016, pp. 3675--3683.

\bibitem{le2018hierarchical}
H.~M. Le, N.~Jiang, A.~Agarwal, M.~Dud{\'\i}k, Y.~Yue, and H.~Daum{\'e}~III,
  ``Hierarchical imitation and reinforcement learning,'' \emph{arXiv preprint
  arXiv:1803.00590}, 2018.

\bibitem{levy2018hierarchical}
A.~Levy, R.~Platt, and K.~Saenko, ``Hierarchical reinforcement learning with
  hindsight,'' \emph{arXiv preprint arXiv:1805.08180}, 2018.

\bibitem{nachum2018data}
O.~Nachum, S.~S. Gu, H.~Lee, and S.~Levine, ``Data-efficient hierarchical
  reinforcement learning,'' in \emph{Advances in Neural Information Processing
  Systems}, 2018, pp. 3307--3317.

\bibitem{nachum2018near}
O.~Nachum, S.~Gu, H.~Lee, and S.~Levine, ``Near-optimal representation learning
  for hierarchical reinforcement learning,'' \emph{arXiv preprint
  arXiv:1810.01257}, 2018.

\bibitem{dwiel2019hierarchical}
Z.~Dwiel, M.~Candadai, M.~J. Phielipp, and A.~K. Bansal, ``Hierarchical policy
  learning is sensitive to goal space design,'' \emph{arXiv preprint
  arXiv:1905.01537}, 2019.

\bibitem{bacon2017option}
P.-L. Bacon, J.~Harb, and D.~Precup, ``The option-critic architecture,'' in
  \emph{Thirty-First AAAI Conference on Artificial Intelligence}, 2017.

\bibitem{wu2018building}
Y.~Wu, Y.~Wu, G.~Gkioxari, and Y.~Tian, ``Building generalizable agents with a
  realistic and rich 3d environment,'' \emph{arXiv preprint arXiv:1801.02209},
  2018.

\bibitem{levy2017hierarchical}
A.~Levy, R.~Platt, and K.~Saenko, ``Hierarchical actor-critic,'' \emph{arXiv
  preprint arXiv:1712.00948}, 2017.

\bibitem{andreas2017modular}
J.~Andreas, D.~Klein, and S.~Levine, ``Modular multitask reinforcement learning
  with policy sketches,'' in \emph{Proceedings of the 34th International
  Conference on Machine Learning}, 2017, pp. 166--175.

\bibitem{sohn2018hierarchical}
S.~Sohn, J.~Oh, and H.~Lee, ``Hierarchical reinforcement learning for zero-shot
  generalization with subtask dependencies,'' in \emph{Advances in Neural
  Information Processing Systems}, 2018, pp. 7156--7166.

\bibitem{das2018neural}
A.~Das, G.~Gkioxari, S.~Lee, D.~Parikh, and D.~Batra, ``Neural modular control
  for embodied question answering,'' \emph{arXiv preprint arXiv:1810.11181},
  2018.

\bibitem{gordon2018iqa}
D.~Gordon, A.~Kembhavi, M.~Rastegari, J.~Redmon, D.~Fox, and A.~Farhadi, ``Iqa:
  Visual question answering in interactive environments,'' in \emph{Proceedings
  of the IEEE Conference on Computer Vision and Pattern Recognition}, 2018, pp.
  4089--4098.

\bibitem{osa2019hierarchical}
T.~Osa, V.~Tangkaratt, and M.~Sugiyama, ``Hierarchical reinforcement learning
  via advantage-weighted information maximization,'' \emph{arXiv preprint
  arXiv:1901.01365}, 2019.

\bibitem{ye2020seeing}
X.~Ye and Y.~Yang, ``From seeing to moving: A survey on learning for visual
  indoor navigation (vin),'' \emph{arXiv preprint arXiv:2002.11310}, 2020.

\bibitem{kulhanek2019vision}
J.~Kulh{\'a}nek, E.~Derner, T.~de~Bruin, and R.~Babu{\v{s}}ka, ``Vision-based
  navigation using deep reinforcement learning,'' in \emph{2019 European
  Conference on Mobile Robots (ECMR)}.\hskip 1em plus 0.5em minus 0.4em\relax
  IEEE, 2019, pp. 1--8.

\bibitem{ye2019gaple}
X.~Ye, Z.~Lin, J.-Y. Lee, J.~Zhang, S.~Zheng, and Y.~Yang, ``Gaple:
  Generalizable approaching policy learning for robotic object searching in
  indoor environment,'' \emph{IEEE Robotics and Automation Letters}, vol.~4,
  no.~4, pp. 4003--4010, 2019.

\bibitem{yang2018visual}
W.~Yang, X.~Wang, A.~Farhadi, A.~Gupta, and R.~Mottaghi, ``Visual semantic
  navigation using scene priors,'' \emph{arXiv preprint arXiv:1810.06543},
  2018.

\bibitem{druon2020visual}
R.~Druon, Y.~Yoshiyasu, A.~Kanezaki, and A.~Watt, ``Visual object search by
  learning spatial context,'' \emph{IEEE Robotics and Automation Letters},
  vol.~5, no.~2, pp. 1279--1286, 2020.

\bibitem{wu2019bayesian}
Y.~Wu, Y.~Wu, A.~Tamar, S.~Russell, G.~Gkioxari, and Y.~Tian, ``Bayesian
  relational memory for semantic visual navigation,'' in \emph{Proceedings of
  the IEEE International Conference on Computer Vision}, 2019, pp. 2769--2779.

\bibitem{anderson2018vision}
P.~Anderson, Q.~Wu, D.~Teney, J.~Bruce, M.~Johnson, N.~S{\"u}nderhauf, I.~Reid,
  S.~Gould, and A.~van~den Hengel, ``Vision-and-language navigation:
  Interpreting visually-grounded navigation instructions in real
  environments,'' in \emph{Proceedings of the IEEE Conference on Computer
  Vision and Pattern Recognition}, 2018, pp. 3674--3683.

\bibitem{sutton2018reinforcement}
R.~S. Sutton and A.~G. Barto, \emph{Reinforcement learning: An
  introduction}.\hskip 1em plus 0.5em minus 0.4em\relax MIT press, 2018.

\bibitem{mnih2016asynchronous}
V.~Mnih, A.~P. Badia, M.~Mirza, A.~Graves, T.~Lillicrap, T.~Harley, D.~Silver,
  and K.~Kavukcuoglu, ``Asynchronous methods for deep reinforcement learning,''
  in \emph{International conference on machine learning}, 2016, pp. 1928--1937.

\bibitem{van2015deep}
H.~Van~Hasselt, A.~Guez, and D.~Silver, ``Deep reinforcement learning with
  double q-learning,'' \emph{arXiv preprint arXiv:1509.06461}, 2015.

\bibitem{anderson2018evaluation}
P.~Anderson, A.~Chang, D.~S. Chaplot, A.~Dosovitskiy, S.~Gupta, V.~Koltun,
  J.~Kosecka, J.~Malik, R.~Mottaghi, M.~Savva \emph{et~al.}, ``On evaluation of
  embodied navigation agents,'' \emph{arXiv preprint arXiv:1807.06757}, 2018.

\bibitem{ye2021GRG}
X.~Ye and Y.~Yang, ``Hierarchical and partially observable goal-driven policy
  learning with goals relational graph,'' in \emph{Proceedings of the IEEE/CVF
  Conference on Computer Vision and Pattern Recognition}, 2021.

\end{thebibliography}
\bibliographystyle{IEEEtran}

\end{document}